\documentclass[conference]{IEEEtran}
\usepackage{times}

\usepackage[numbers]{natbib}
\usepackage{multicol}
\usepackage[bookmarks=true]{hyperref}
\usepackage{multirow}

\usepackage{hyperref}
\usepackage{url}

\usepackage{mathtools}
\usepackage{amsfonts}  %
\usepackage{graphics}
\usepackage[pdftex]{graphicx}
\usepackage{wrapfig}
\usepackage{caption}
\usepackage{color}
\usepackage{dsfont}
\usepackage[]{mdframed}
\usepackage{algorithm}
\usepackage{algorithmic}
\usepackage{xparse}
\usepackage{amsmath}
\usepackage{mathtools}
\usepackage{amssymb}
\usepackage{amsthm}
\usepackage{amssymb}

\usepackage{booktabs}

\usepackage{xcolor} %
\usepackage{hyperref} %
\hypersetup{
    colorlinks=true,
}

\DeclarePairedDelimiterX{\infdivx}[2]{(}{)}{%
  #1\;\delimsize\|\;#2%
}

\usepackage{expl3}
\ExplSyntaxOn
\newcommand\latinabbrev[1]{
  \peek_meaning:NTF . {%
    #1\@}%
  { \peek_catcode:NTF a {%
      #1.\@ }%
    {#1.\@}}}
\ExplSyntaxOff
\def\eg{\latinabbrev{e.g}}

\def\etc{\latinabbrev{etc}}

\NewDocumentCommand \proposition {g g g g} {\texttt{#1}(#2
  \IfValueTF{#3}{,\,#3}{}
  \IfValueTF{#4}{,\,#4}{}
  )
}

\NewDocumentCommand \actioncall {g g g g} {\text{#1}(#2
  \IfValueTF{#3}{,#3}{}
  \IfValueTF{#4}{,#4}{}
  \texttt{)}
}

\providecommand{\rebuttal}[1]{{\color{black} #1}}

\def \MethodName {Marking Open-world Keypoint Affordances }
\def \MethodAcronym {MOKA}
\IEEEoverridecommandlockouts

\begin{document}

\title{
\vspace{-0.1cm}
\MethodAcronym: Open-World Robotic Manipulation through Mark-Based Visual Prompting
\vspace{-0.1cm}
}

\author{
    Fangchen Liu\textsuperscript{*1} \quad
    Kuan Fang\textsuperscript{*1} \quad
    Pieter Abbeel\textsuperscript{1} \quad
    Sergey Levine\textsuperscript{1}
    \thanks{$^{*}$Equal contribution. $^{1}$University of California, Berkeley. Correspondance to: \texttt{\{fangchenliu, kuanfang\}@berkeley.edu}}
    \vspace{0.4cm} \\
    \url{https://moka-manipulation.github.io}
    \vspace{-0.2cm}
    }

\maketitle

\begin{abstract}
Open-world generalization requires robotic systems to have a profound understanding of the physical world and the user command to solve diverse and complex tasks. While the recent advancement in vision-language models (VLMs) has offered unprecedented opportunities to solve open-world problems, how to leverage their capabilities to control robots remains a grand challenge. In this paper, we introduce \MethodName(\MethodAcronym), an approach that employs VLMs to solve robotic manipulation tasks specified by free-form language instructions. Central to our approach is a compact point-based representation of affordance, which bridges the VLM's predictions on observed images and the robot's actions in the physical world. By prompting the pre-trained VLM, our approach utilizes the VLM's commonsense knowledge and concept understanding acquired from broad data sources to predict affordances and generate motions. To facilitate the VLM's reasoning in zero-shot and few-shot manners, we propose a visual prompting technique that annotates marks on images, converting affordance reasoning into a series of visual question-answering problems that are solvable by the VLM. We further explore methods to enhance performance with robot experiences collected by \MethodAcronym~through in-context learning and policy distillation. We evaluate and analyze \MethodAcronym's performance on various table-top manipulation tasks including tool use, deformable body manipulation, and object rearrangement.
\end{abstract}

\IEEEpeerreviewmaketitle

\begin{figure}[ht]
    \centering
    \includegraphics[width=0.5\textwidth]{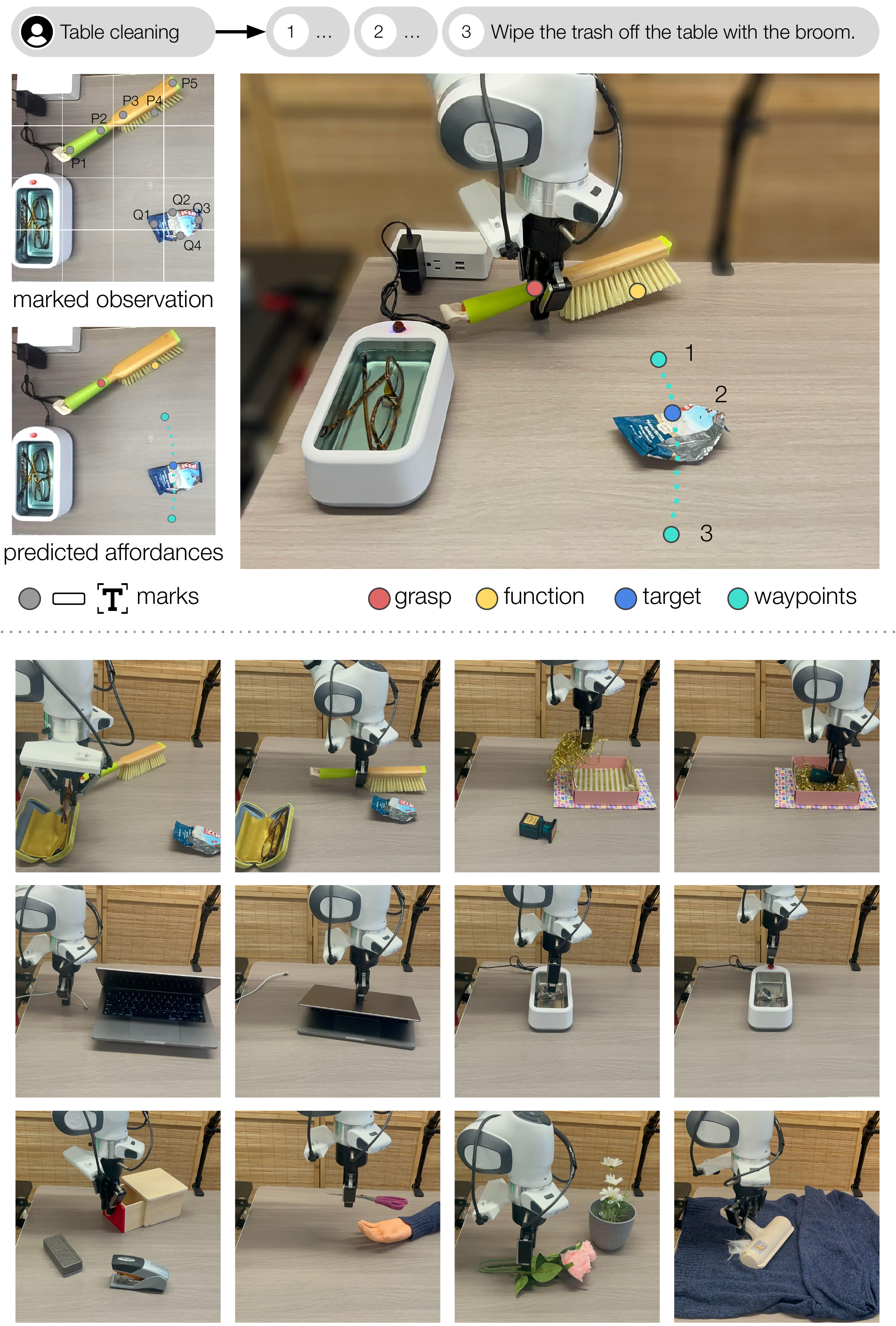} 
    \caption{To solve manipulation tasks with unseen objects and goals, \MethodAcronym~employs a VLM to generate motions through a point-based affordance representation (plotted as colorful dots on the images). By annotating marks (\eg, candidate points, grids, and captions) on the observed 2D image, \MethodAcronym~converts motion generation into a series of visual question-answering problems that the VLM can solve. }
    \label{fig:intro}
    \vspace{-0.1cm}
\end{figure}
\section{Introduction}

The pursuit of open-world generalization poses a significant challenge for robotic systems: Solving various tasks commanded by humans in complex and diverse environments requires a profound understanding of the physical world.
An appealing prospect for handling this challenge is to employ large pre-trained models that encapsulate extensive prior knowledge from broad data and bringing it to bear on novel problems. 
Recent advances in large language models (LLMs) and vision-language models (VLMs) provide particularly promising tools in this regard, with their emergent and fast-growing conceptual understanding, commonsense knowledge, and reasoning abilities~\citep{devlin2018bert,radford2018improving, radford2019language, brown2020language, chowdhery2023palm, achiam2023gpt, radford2021learning, li2022blip, ramesh2021zero, li2023blip}.  
However, existing large models pre-trained on Internet-scale data still lack the capabilities to understand 3D space, contact physics, and robotic control, not to mention the knowledge about the embodiment and environment dynamics in each specific scenario.
This creates a large gap between the promising trend in computer vision and natural language processing and applying them to robotics.
It remains an open question how such tools can guide a robotic system to solve manipulation tasks in the physical world. 

Recently, a growing body of research has been dedicated to utilizing pre-trained large-scale models for robotic control. 
To incorporate broad knowledge, these approaches either directly prompt or fine-tune the large models to generate plans~\cite{ahn2022can, huang2022inner, huang2022language, chen2023open}, rewards~\cite{ma2023eureka, kwon2023reward, mahmoudieh2022zero, yu2023language}, codes~\cite{liang2023code, singh2023progprompt, wang2023voyager}, \etc.
Despite the encouraging results they have demonstrated, these approaches are subject to notable limitations.
Since the advances in LLMs precede VLMs, many previous approaches first process the raw sensory inputs to obtain the language description of the environment and then query LLMs to perform reasoning and planning in the language domain. However, relying solely on high-level language descriptions may overlook the nuanced visual details of environments and objects, which are vital for accurately completing tasks. Moreover, existing approaches usually require non-trivial effort in designing in-context examples~\cite{huang2023voxposer, liang2023code, singh2023progprompt} to ensure LLMs can produce desired predictions on similar tasks. As a result, the tasks that can be solved by these approaches are largely constrained by such manual efforts.

In this work, we study how to effectively endow robots with the ability to solve novel manipulation tasks specified by free-form language instructions using VLMs. Our key insight is to find an intermediate affordance representation that connects the VLM's prediction on images with the robot's motion in the physical world. This affordance representation should satisfy two essential requirements. First, it should be feasible for the VLM to predict such representation based on the visual observation of the environment and the task description. Second, the representation should compactly capture the information about the relevant properties of the robot's motion so that it can be easily executed on the robot.  

To this end, we propose \MethodName(\MethodAcronym), an approach that employs VLMs for robotic manipulation through mark-based visual prompting.
As shown in Fig.~\ref{fig:intro}, \MethodAcronym~leverages a compact affordance representation consisting of a set of keypoints and waypoints, defined on open sets of objects and tasks. 
This point-based affordance representation is then used to specify the desired motion for the robot to solve the task.
To generate the motions given the free-form language descriptions, \MethodAcronym~uses hierarchical visual prompting to convert the affordance reasoning problem into a series of visual question-answering problems.
Drawing inspirations from recent advances in visual question-answering~\citep{yang2023setofmark}, we use mark-based visual prompting to enable the VLM to attend to the important visual cues in the observation image and further simplify the point generaton problem into multiple choice questions. 
As shown in the top-left part of Fig.~\ref{fig:intro}, we plot candidate points on the image and query the VLM to select the optimal set of points that will result in the desired motion. 
The predicted keypoints and waypoints are used to specify a motion trajectory  which can cover a wide range of manipulation skills such as picking, placing, pressing, tool use, \etc. 

We demonstrate the effectiveness of \MethodAcronym~on various manipulation tasks, with robustness across the variations of instructions, objects, and initial arrangements. We further use \MethodAcronym~to collect successful trajectories in each task. The collected trajectories can be used as in-context examples to bootstrap the performance of VLM, and can also be leveraged as demonstrations to train a better student policy. Our experiments show that \MethodAcronym~can achieve state-of-the-art performance on various evaluation tasks in a zero-shot manner, with consistent improvements using clean and intuitive in-context examples. To summarize, our contributions are:
\begin{itemize}
    \item We introduce a point-based affordance representation that bridges the VLM's prediction on RGB images and the robot's motion in the physical world.
    \item We propose a mark-based visual prompting approach that converts affordance reasoning into a series of visual question-answering problems.
    \item We demonstrate that the proposed approach, \MethodAcronym, can effectively generate motions for open-world manipulation in zero-shot and few-shot manners.
\end{itemize}

\section{Related Work}

\begin{figure*}[t]
    \centering
    \includegraphics[width=\textwidth]{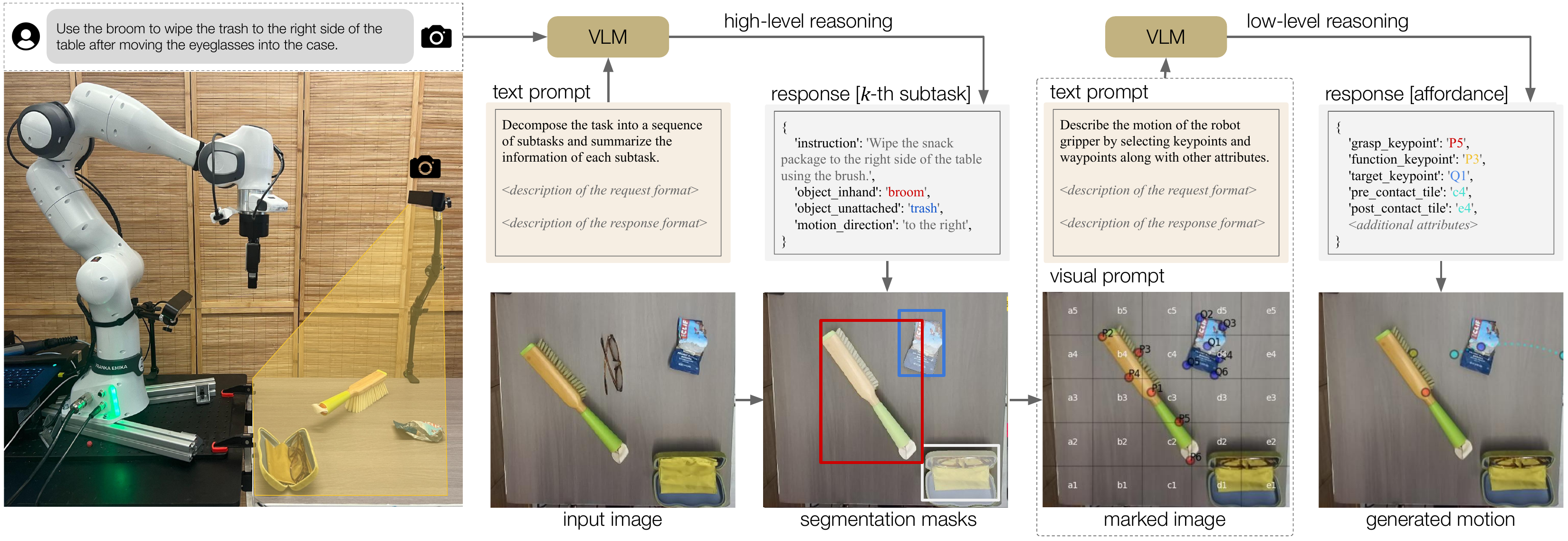} 
    \caption{\textbf{Overview of \MethodAcronym.} We propose a hierarchical approach to prompt the VLM to perform affordance reasoning based on the free-form langauage instruction of the task and the visual observation of the environment. On the high level, we query the VLM to decompose the free-form language description of the task into a sequence of subtasks and summarize the subtask information. On the low level, the VLM is prompted to produce the keypoints and additional attributes for the affordance representation defined in Sec.~\ref{sec:motion_primitive}. 
    }
    \vspace{-0.5cm}
    \label{fig:method}
\end{figure*}

\textbf{Large language models and vision-language models. }
Recent advances in several domains and applications have been greatly influenced by the substantial progress achieved through large language models (LLMs) and vision language models (VLMs)~\citep{devlin2018bert,radford2018improving, radford2019language, brown2020language, chowdhery2023palm, achiam2023gpt, radford2021learning, li2022blip, ramesh2021zero, li2023blip}. While these models can already solve various tasks in a zero-shot manner~\citep{achiam2023gpt}, well-designed prompts still serve an important role in further eliciting more advanced capabilities.
As demonstrated by~\citet{brown2020language}, few-shot prompting can match or surpass the performance of fine-tuning on LLMs. Additionally, other prompting techniques~\citep{wei2022chain, zhou2022least, kojima2022large, yao2023tree} have been proposed to improve LLMs' reasoning capabilities. Although these methodologies may initially appear enigmatic, they have been empirically validated to consistently demonstrate scalability across various models~\citep{brown2020language, chowdhery2023palm, achiam2023gpt, yang2023dawn}. Apart from the language prompts, recent vision models~\citep{kirillov2023segment, zou2023segment, li2023semantic} are capable of supporting visual prompts, including points, boxes, masks, and texts. Such visual prompts exhibit greater diversity in both form and content, harnessing vision-language models for various application scenarios like perception~\citep{yang2023fine}, image editing~\citep{shi2023dragdiffusion} and reasoning~\citep{yang2023setofmark}. 

Incorporating robust visual reasoning capabilities becomes imperative to build autonomous robots capable of making decisions in unstructured environments. Although the predictions by existing VLMs cannot seamlessly engage in zero-shot reasoning, they can still be effectively harnessed across diverse tasks. Prior work such as SoM~\citep{yang2023setofmark} draws visual marks on the objects in an image using numbers and segmentation masks and demonstrates the mark-based visual prompting scheme unleashes the reasoning capabilities of GPT-4V, such as object counting and inferring spatial relationship. Inspired by SoM~\citep{yang2023setofmark}, we wleverage GPT-4V for robotic manipulation. We represent manipulation skills with a set of affordance points and motion waypoints. Instead of using object segmentation masks, we use keypoints and grid cells as visual prompts as shown in Fig.~\ref{fig:intro}, and then query GPT-4V to choose the affordance points and motion waypoints from the visual marks, followed by executable point-based motion plans. 

\textbf{Foundation models in robotics.}
Building upon the successes of foundation models, the field of robotics is currently witnessing a rising interest in utilizing LLMs and other foundation models in various applications.
With large model capacities, transformers~\cite{vaswani2023attention} and VLMs can be trained from scratch or fine-tuned to directly predict actions through imitation learning and reinforcement learning~\cite{fang2019smt, brohan2023rt, jangBCZZeroShotTask2021, lynchLanguageConditionedImitation2021, grif2023}. However, training large models for open-world scenarios would usually require extensive data and supervision.
Pre-trained LLMs can also be used to generate high-level task plan in natural language~\citep{ahn2022can, huang2022inner, huang2022language, chen2023open}, executable programs~\citep{liang2023code, singh2023progprompt, wang2023voyager} or value function~\cite{huang2023voxposer, lin2023text2motion} for low-level behaviours, environmental reward and feedback for reinforcement learning~\citep{ma2023eureka, kwon2023reward, mahmoudieh2022zero, yu2023language}. However, many of these approaches necessitate the conversion of both the task and the observations into the textual form. While this can be easily accomplished in simulated environments with ground-truth object states, tasks in the real world require the utilization of robust and precise perception modules. For performing open-ended navigation and manipulation in the real world~\citep{stone2023open, chen2023open, gadre2023cows}, open-world VLMs~\citep{kirillov2023segment, zou2023segment, li2023semantic, liu2023grounding, minderer2022simple, li2022grounded} are often used to extract visual scene information before converting the observation to textual form. However, the process of converting an image into text may result in the loss of important details, such as shape and geometric information. With the recent advancements in VLMs, it is now conceivable that the role of state estimation combined with language models can be replaced by a single, powerful VLM, such as GPT-4V, which is leveraged by \MethodAcronym. ~\citet{hu2023look} utilizes GPT-4V to get semantic language plans and convert each language plan to a set of pre-defined low-level skills. Different from~\citet{hu2023look}, we represent affordances and motions using a set of points, and query GPT-4V to select the corresponding points (e.g. grasping point, function point) in each task, which can be directly translated to low-level point-based motions. While preserving the visual reasoning capabilities of VLMs, \MethodAcronym~provides a framework with more general and flexible low-level motion, which doesn't require prior knowledge about a specific task compared to pre-defined skills.

\textbf{Affordance reasoning for robotic control.}
Affordance~\citep{gibson2014ecological} refers to the ability to perform a certain action with an object in a given environment. Much of the research on affordances focuses on understanding interactions with objects, as demonstrated by~\citep{sun2010learning, hermans2011affordance, aldoma2012supervised}. In the field of robot manipulation, keypoints and other representation forms are often used to provide compact information about the environment and the objects~\citep{choi2010real, finn2017deep, van2010gravity, maitin2010cloth, miller2011parametrized, fang2018demo2vec, Manuelli2019kPAMKA, qin2019keto, bahl2023affordances, jiang2021giga, Khazatsky2021WhatCI, fang2022planning, fang2022flap}, representing the affordance in a structured way. Among these works, the most related one to ours is KETO~\citep{qin2019keto}, which predicts affordance and functional keypoints on the tool objects by learning from interactions with a trajectory optimization formulation as in \citet{Manuelli2019kPAMKA}. In contrast to KETO~\citep{qin2019keto}, our keypoints selection procedure doesn't require any model training. We design an automated process to annotate keypoints as visual marks on a 2D image and leverage the broad knowledge from GPT-4V to select affordance and motion keypoints for manipulation.

\section{Problem Statement}
\label{sec:problem_formulation}

Our goal is to enable robots to perform manipulation tasks involving unseen objects and goals. 
Each task is described by a free-form language instruction $l$.
To achieve the success, the robot needs to interact with the environment across one or multiple stages.
An example task is shown in Fig.~\ref{fig:intro}, in which the robot is commanded to clean the table by the instruction ``Wipe the trash off the table after moving away the eyeglasses.''.

The physical interaction at each stage is referred to as a \textit{subtask}.
Each subtask can be an interaction with objects in hand (\eg, lifting up an object, opening a drawer, turning the faucet), an interaction with environmental objects unattached to the robot (\eg, pushing an obstacle, pressing a button), and tool use involving grasping a tool object to make contact with another object (\eg, poking with a stick, sweeping with a foam, cutting with a knife).
In this example above, the robot needs to perform two subtasks: First, put the eyeglasses between the broom and the snack package back into the case; and second, use the broom to sweep the trash. 
Since we do not assume the sequence of subtasks to be given beforehand, the robot needs to decompose the task into a sequence of feasible subtasks based on the free-form language instruction $l$ and solve each subtask sequentially.

We consider a table-top setting with a robotic arm and RGBD cameras, as shown in Fig.~\ref{fig:method}. 
At each time step $t$, an observation $s_t$ is received from the environment and an action $a_t$ is selected to command the robot.
The observation $s_t$ consists of the RGBD images captured by the camera sensors as well as the proprioception of the robot. 
The action $a_t$ is defined as the 6-DoF gripper pose and the finger status.

In this paper, we aim to solve such tasks by leveraging vision-language models (VLMs) with an effective visual prompting strategy. We use $\mathcal{M}$ to denote the VLM, which can take a list of language and visual inputs in a specific order, and generate text responses accordingly. The responses of $\mathcal{M}$ can be controlled by designing different input prompts, which specify information such as the problem descriptions, the input and output formats, and in-context examples~\cite{huang2023voxposer, liang2023code, ma2023eureka, singh2023progprompt}. Specifically, we design text and visual prompts to enable $\mathcal{M}$ to generate desired text responses in a structured format, which can be parsed into point coordinates and other attributes to characterize the robot motion for the downstream usage.

\section{\MethodName}

We propose \MethodName~(\MethodAcronym), an approach that leverages the emergent reasoning capability of Vision-Language Models (VLMs) to guide a motion planner to solve unseen tasks specified by free-form language instructions (see Fig.~\ref{fig:method}). 
In this section, we will start by introducing the point-based affordance representations and how they are used for generating motions.
Then, we will describe a hierarchical framework that effectively prompts VLMs for affordance reasoning. 
After this, we will explain a novel visual prompting approach that lifts the VLM's reasoning capabilities by annotating a set of marks, including candidate points, grids, and labels, on 2D images.
Lastly, we will investigate two ways to bootstrap the performance of \MethodAcronym by utilizing successful experiences collected in the physical world. 

\subsection{Point-Based Affordance Representations}
\label{sec:motion_primitive}

To leverage VLMs for solving open-world manipulation tasks, there needs to be an interface that connects the predictions by the VLM and the motions performed by the robot. 
To achieve this goal, we design an affordance representation defined on 2D images. 
Produced as the end result by the VLM, the affordance representation specifies the desired motion.

By extending the definitions in \citet{Manuelli2019kPAMKA} and \citet{qin2019keto}, we design a point-based affordance representation that can cover a wide range of manipulation tasks.
Instead of separately devising motion primitives for different pre-defined skills, we use a unified set of keypoints and waypoints to specify the motion. 
These points are predicted by VLMs on 2D images and converted to poses in the $SE(3)$ space. 
Then a smooth motion trajectory is generated based on these poses. 
To perform the task, the robot interacts with the environment by following the generated motion trajectory.

We specify the robot's motion in an object-centric manner, as shown in Fig.~\ref{fig:method}. Following the discussion in Sec.~\ref{sec:problem_formulation}, we would like this representation to apply to different types of interactions with objects in the environment. Therefore, we consider two types of objects, $o_\text{in-hand}$ (\eg, the broom) and $o_\text{unattached}$ (\eg, the trash), and specify the motion with a grasping phase and a manipulation phase. 
In the grasping phase, the robot reaches and grasps an object $o_\text{in-hand}$ from the environment.
Then in the manipulation phase, the robot performs a motion and makes contact with another object $o_\text{unattached}$, either directly or using $o_\text{in-hand}$ as a tool. 
In some scenarios, only one of these two types of objects is interacted with by the robot, either $o_\text{in-hand}$ (\eg, unplugging a cable, opening a drawer) or $o_\text{unattached}$ (\eg, pressing a button), and one of the two phases can be skipped accordingly.

We now describe the definition of the keypoints and waypoints as well as how they are used to specify the motions in both phases.
These points are illustrated in Fig.~\ref{fig:method}, and more examples can be found in Sec.~\ref{sec:experiments_qulitative}.
Following the practice of \citet{Manuelli2019kPAMKA} and \citet{qin2019keto}, we use the \textbf{grasping keypoint} $x_\text{grasp}$ to specify the position on $o_\text{in-hand}$ where the robot gripper should hold the object.
If $o_\text{in-hand}$ is not involved in a task, the grasping phase will be skipped. 
For the manipulation phase, the robot's gripper follows a motion trajectory specified by an additional set of points.  
The \textbf{function keypoint} $x_\text{function}$ specifies the part of $o_\text{in-hand}$ that will make contact with $o_\text{unattached}$ in the manipulation phase.
If $o_\text{in-hand}$ is not specified, $x_\text{function}$ will be on the robot gripper, and the contact will directly be made between the robot and $o_\text{unattached}$.
Correspondingly, the \textbf{target keypoint} $x_\text{target}$ is the part of $o_\text{unattached}$ that will be contacted by $x_\text{function}$ during the manipulation phase.
We also introduce the \textbf{pre-contact waypoints} $x_\text{pre-contact}$ and the \textbf{post-contact waypoints} $x_\text{post-contact}$ defined in free space, which dictates the manipulation motion along with the keypoints defined on the objects. 

During the manipulation phase, the robot moves the gripper such that $x_\text{function}$ follows the path that sequentially connects the $x_\text{pre-contact}$, $x_\text{target}$, and $x_\text{post-contact}$. 
Besides following the path, we also require the robot gripper to follow the specified \textbf{grasping orientation} $R_\text{grasp}$ and \textbf{manipulation orientation} $R_\text{manipulate}$ during the two phases, respectively. 
To better illustrate the design of our point-based motion, we provide examples of the predicted point specifications and the resultant motions from our experiments in Sec.~\ref{sec:experiments_qulitative} and the Appendix.
In \MethodAcronym, this set of keypoints and \textbf{additional attributes} (described in Sec.~\ref{sec:visual_prompting}) are summarized in a dictionary as the affordance representation (see Fig.~\ref{fig:method}). 

\subsection{Affordance Reasoning with Vision-Language Models}
\label{sec:hierarchical_prompting}

To predict the defined affordance representations, we employ the VLM $\mathcal{M}(\cdot)$ (defined in Sec.~\ref{sec:problem_formulation}), which is pre-trained on Internet-scale data for solving general visual question-answering (VQA) problems.
Using a hierarchical prompting framework as shown in Fig.~\ref{fig:method}, \MethodAcronym~converts affordance reasoning into a series of VQA problems that are solvable by the pre-trained VLM.

The hierarchical prompting framework takes as input the free-form language description $l$ of the task and an RGB image observation of the environment $s_t$.
\MethodAcronym~examines the initial observation $s_0$ and decomposes the task $l$ into a sequence of subtasks using the VLM.
For each of the subtasks, the VLM is asked to provide the summary of the subtask instruction, the description of the corresponding $o_\text{in-hand}$, $o_\text{unattached}$, as well as the description of the motion (\eg, ``from left to right'').
On the low level, given the response from the high-level reasoning and the visual observation $s_{t(k)}$ at the beginning of $k$-th subtask (at the time step $t(k)$), the VLM is queried again with a different prompt to produce the affordance representation defined in Sec.~\ref{sec:motion_primitive}.
In the remainder of this section, we will describe the input formats, output formats, and prompt designs that we use to instantiate this method.
Further details can be found in the Appendix.

\textbf{High-level reasoning.} 
Given the initial observation $s_0$ and the language description $l$, we first query the VLM $\mathcal{M}$ with the language prompt $\rho_\text{high}$ to produce the response $y_\text{high}$: 
\begin{equation}
    y_\text{high} = \mathcal{M}([\rho_\text{high}, l, s_0]).
\end{equation}
The representation $y_\text{high}$ is a string that contains structured information for the $K$ subtasks that are needed to solve the task.
We design the prompt so as to require the VLM to produce $y_\text{high}$ as a list of dictionaries.
As shown in Fig.~\ref{fig:method}, each dictionary contains the language description of a subtask (\eg, \textit{``Wipe the snack package to the right side of the table using the broom.''}), as well as detailed information to facilitate motion generation, including the description of $o_\text{in-hand}$ (\eg, \textit{``broom''}), the object name of $o_\text{unattached}$ (\eg, \textit{``snack package''}), and the description of the motion (\eg, \textit{``from left to right''}). 
This high-level plan will be used as an intermediate result for producing the detailed affordance representation through the low-level reasoning with the VLM. 
\rebuttal{
The prompt $\rho_\text{high}$ used for high-level reasoning contains the description of the request (the input instruction and the annotated image) and the description of the response format.
The complete high-level reasoning prompts are described in TABLE~\ref{tab: input-description}.
} 

\textbf{Low-level reasoning.} 
Next, we prompt the VLM once again to produce the affordance representation defined in Sec.~\ref{sec:motion_primitive} as $y_{\text{low}}^k$, conditioning on the high-level representation $y_\text{high}$ and the visual observation $s_{t(k)}$ at the beginning of the $k$-th subtask. 
Instead of directly predicting 3D coordinates on 2D images, which is challenging and even ill-defined, we query the VLM to output 2D coordinates on the images and deproject them back to the 3D space.
The three keypoints $x_{\text{grasp}}, x_{\text{function}}$ and $x_{\text{target}}$ are defined on the object surface, and thus we can compute the 3D coordinates using the corresponding depth value of the 2D location based on the RGB image and camera parameters.
For the waypoints in free space, we query the VLM to predict the desired height in the text.
To produce such an affordance representation $y_{\text{low}}^k$, we query the VLM again by
\begin{equation}
    y_{\text{low}}^k = \mathcal{M}([\rho_\text{low}, y_{\text{high}}^k , f(s_{t(k)})]),
\end{equation}
where $y_{\text{high}}^k$ is the substring corresponding to the $k$-th subtask extracted from $y_\text{high}$, and $f(\cdot)$ is a function that process the raw visual observation $s_{t(k)}$.
\rebuttal{
The prompt $\rho_\text{low}$ used for low-level reasoning contains the description of the request (the response format of the high-level reasoning and the annotated image) and the description of the response (explanation of the point-based affordance representation and the desired format of the response).
The complete low-level reasoning prompts are described in TABLE~\ref{tab: input-description}, TABLE~\ref{tab:explanation}, and TABLE~\ref{tab:motion-output}.
}

\subsection{Mark-Based Visual Prompting}
\label{sec:visual_prompting}

To perform the low-level reasoning mentioned in the previous section, we need the VLM to generate keypoints and waypoints on 2D images in order to execute a specific motion for a subtask. Since VLMs are better at multiple-choice problems than directly producing continuous-valued locations, we employ a mark-based visual prompting strategy to extract the desired output from VLMs, which we will describe in this subsection.

Inspired by \citet{Yang2023SetofMarkPU}, \MethodAcronym~uses a set of marks as visual prompts to enable VLM to apply its reasoning capability to predict the point-based affordance representation as shown in Fig.~\ref{fig:method}. 
Consisting of dots, grids, and text notations annotated on the visual observation, these marks play an important role in the reasoning process.
Proposed by open-world object detection and segmentation algorithms, these marks facilitate visual reasoning by encouraging the VLM to attend to the target objects and other task-relevant information in the image.
We annotate marks as candidate parts and regions for the VLM to choose the points from, converting the original problem of directly generating coordinates into multiple-choice questions, which is usually more tractable for existing VLMs.

To select the keypoints, which are defined on the in-hand object $o_\text{in-hand}$ and the unattached object $o_\text{unattached}$ suggested by the high-level reasoning in Sec.~\ref{sec:hierarchical_prompting}, we propose and plot candidate keypoints on these objects. 
Given the names of $o_\text{in-hand}$ and $o_\text{unattached}$, we first segment these two objects using GroundedSAM~\cite{ren2024grounded}, which combines GroundingDINO~\cite{liu2023grounding} and SAM~\cite{zou2023segment} to extract segmentation masks of objects specified by a text prompt. After we obtain the segmentation masks of $o_\text{in-hand}$ and $o_\text{unattached}$, we perform farthest point sampling~\citep{qi2017pointnet} on the object contour to obtain $K$ boundary points. Together with its geometric center, and overlay the $K+1$ candidate keypoints on each object. 
Each candidate keypoint is assigned an index, which is annotated next to it as a reference. 
To avoid confusion, we use different colors for candidate keypoints on $o_\text{in-hand}$ and $o_\text{unattached}$ and use the caption in the format of $P_i$ and $Q_j$ respectively, where $i$ and $j$ are integers. More implementation details can be found in Appendix~\ref{sec:keypoint-implementation}.

Selecting waypoints in free space involves searching over a much larger region. 
Instead of directly sampling points in the entire workspace, we divide the observed RGB image into an $m \times n$ grid, where $m$ and $n$ are integers. Both $m$ and $n$ are set to $5$ for our evaluation tasks.
The VLM is prompted to choose the tiles in which the pre-contact and post-contact keypoints are supposed to locate in, and then the exact waypoints are sampled uniformly within the tile.
For this purpose, we overlay the grid along with the name of each tile on the image. 
The tile names follow chess notation, which uses letters to specify the columns and integers for the rows. 

\subsection{Motion Generation with Predicted Affordances}
After obtaining the selected keypoints and waypoints from the VLM, we convert the affordance to an executable motion on a real robot. To achieve this goal, we need to lift up all the points from the 2D image to the $SE(3)$ space. 

For the keypoints defined on the object, we can directly take the corresponding point on the registered depth image and transform it into the robot's frame. For the waypoints that are selected from the free space, since they are not attached to any objects, their height needs to be specified in order to be deprojected to 3D space. For this purpose, we also ask the VLM to determine the height of waypoints. We only consider the cases where the waypoints are at the same height as the target point (\eg,  pushing) or above the target point (\eg,  placing) in most common tabletop manipulation scenarios. Additionally, we also prompt the VLM to return the orientation of $o_\text{in-hand}$ during the manipulation phase, which would usually affect the success of tool use tasks. 

Drawing inspirations from \citet{Manuelli2019kPAMKA}, we use the vector from $x_\text{grasp}$ to $x_\text{function}$ to specify the orientation of the object and ask the VLM to predict the orientation from a finite set of options (\eg,  forward, backward, upside, downside, left, right).

Since robust grasping relies on contact physics and gripper design, which is beyond the capability of existing pre-trained VLMs, we combine the VLM predictions with analytical approaches in robotic grasping pipelines. Similar to \citet{Manuelli2019kPAMKA}, we use a grasp sampler to propose candidate grasps based on local geometric information from the observed point cloud. Instead of directly relying on the predicted $x_\text{grasp}$, we use the position and orientation of the grasp candidate that is closest to $x_\text{grasp}$. More implementation details can be found in the Appendix.

\subsection{Bootstrapping through Physical Interactions}

We consider two ways to bootstrap the performance of \MethodAcronym~through physical interactions with the real world.
In both ways, we unroll the VLM policy for the target task to collect robot experiences. 
The success labels of the collected experiences will be annotated by the VLM or humans.

\textbf{In-context learning. }
We use a handful of successful trajectories as in-context examples \rebuttal{(as shown in Fig.~\ref{fig:context})} to guide the high-level and low-level reasoning of the VLM. As discussed in prior work~\cite{brown2020language}, two to three such in-context examples can significantly boost the performance of the VLM.
Without changing the prompts $\rho_\text{high}$ and $\rho_\text{low}$, we append the prompt with three pairs of annotated images and VLM responses from previous successful rollouts by the VLM policy.  

\textbf{Policy distillation. }
\rebuttal{\MethodAcronym~can be used for collecting demonstration datasets for real-world robot learning. We record the multi-view images and robot proprioception states when collecting successful trajectories and train a visuomotor policy using a simple behavior cloning objective. More details can be found in Sec.~\ref{sec:experiments_setup}.}

\section{Experiments}
The primary goal of our experiments is to validate and analyze the behavior of \MethodAcronym~on open-world tasks with a wide range of objects and skills. We design our experiments to investigate the following questions:
\begin{itemize}
    \item Can \MethodAcronym~ effectively reason about affordances based on the images for unseen tasks?
    \item After translating the prediction of \MethodAcronym~into low-level motion, how well does it perform on the task we proposed?
    \item Can \MethodAcronym~improve from real-world trials, either via in-context learning or policy distillation?
\end{itemize}

We first evaluate \MethodAcronym's performance against baseline methods on 4 manipulation tasks in both zero-shot and in-context learning settings. Then we provide qualitative and robustness analysis in Sec.~\ref{sec:experiments_qulitative}. 

Each of our evaluation tasks has two stages, and each task involves a variety of object interaction and tool-use scenarios. A summary of our evaluation tasks can be found in TABLE~\ref{tab:task_desion}. We also tested additional open-world tasks to demonstrate our method, which can be found in Appendix~\ref{sec:additional_results}. 

\begin{table*}[ht]
    \centering
    \begin{tabular}{lcccccccc}  
    \toprule
    & \multicolumn{2}{c}{Table Wiping} & \multicolumn{2}{c}{Watch Cleaning} & \multicolumn{2}{c}{Gift Preparation} & \multicolumn{2}{c}{Laptop Packing} \\
    \cmidrule(lr){2-3} \cmidrule(lr){4-5} \cmidrule(lr){6-7} \cmidrule(lr){8-9}\\
    & Subtask I & Subtask II & Subtask I & Subtask II & Subtask I & Subtask II & Subtask I & Subtask II  \\
    Methods & \\
    \midrule
    Code-as-Policies~\citep{liang2023code}                    & 0.7           & 0.6 & 0.6 & 1.0 & 1.0 & 0.7 & 0.4 & 0.8 \\
    VoxPoser~\citep{huang2023voxposer}                    & 0.6           & 0 & 0.6 & 0.8 & 1.0 & 0.6 & 0.5 & 0.8 \\
    \hline
    MOKA Zero-Shot (Ours)     & 0.6           & 0.6 & 0.7 & 1.0 & 1.0 & 0.7 & 0.5 & 0.8 \\
    MOKA Distilled (Ours)     & \textbf{1.0}  & 0.7 & 0.8 & 0.8 & 1.0 & 0.7 & 1.0 & \textbf{1.0} \\
    MOKA In-Context (Ours)    & 0.9           & \textbf{0.9} & \textbf{0.9} & \textbf{1.0} & \textbf{1.0} & \textbf{0.9} & \textbf{1.0} & 0.9 \\
    \bottomrule
    \end{tabular}
    \caption{Success rate of our method and baselines. Across four tasks, each consists of 2 subtasks, \MethodAcronym~consistently achieves superior performances. We also demonstrated that the performance can be further bootstrapped through physical interactions in the environment using policy distillation and in-context learning.}
    \label{tab:experiment}
\end{table*}

\begin{figure*}[t]
    \centering
    \includegraphics[width=\textwidth]{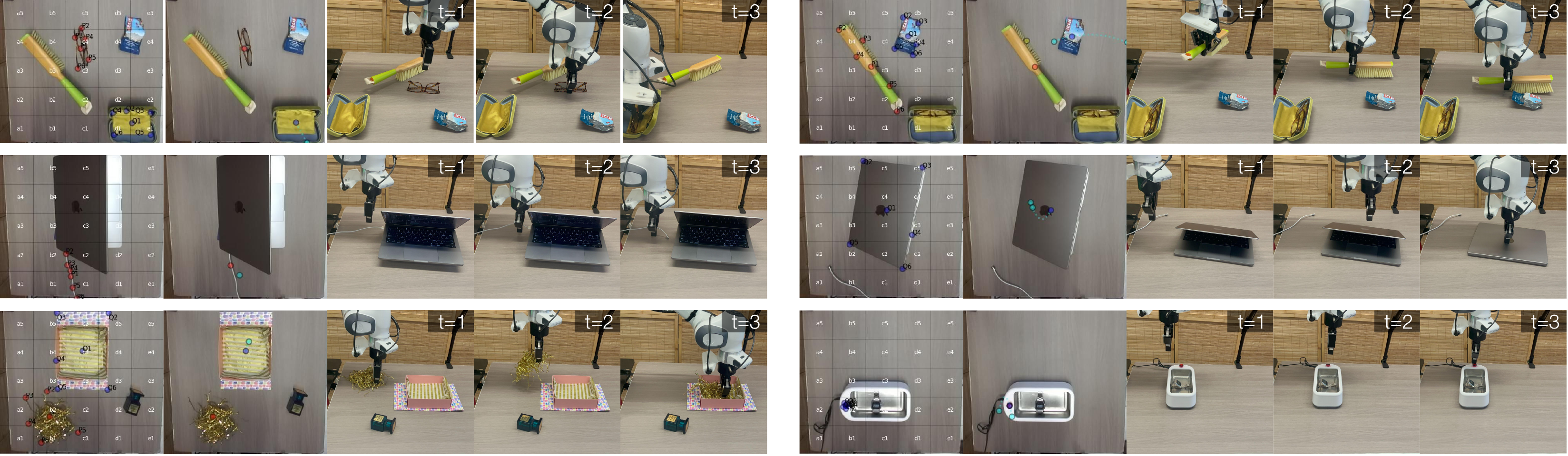} 
    \caption{Example results of the predicted keypoints and motions. On each row, two individual subtasks are displayed. For each subtask, we plot the marked image, the predicted point-based affordances, and three keyframes in the trajectory.}
    \label{fig:example_results}
\end{figure*}
\begin{figure}[t]
    \centering
    \includegraphics[width=0.47\textwidth]{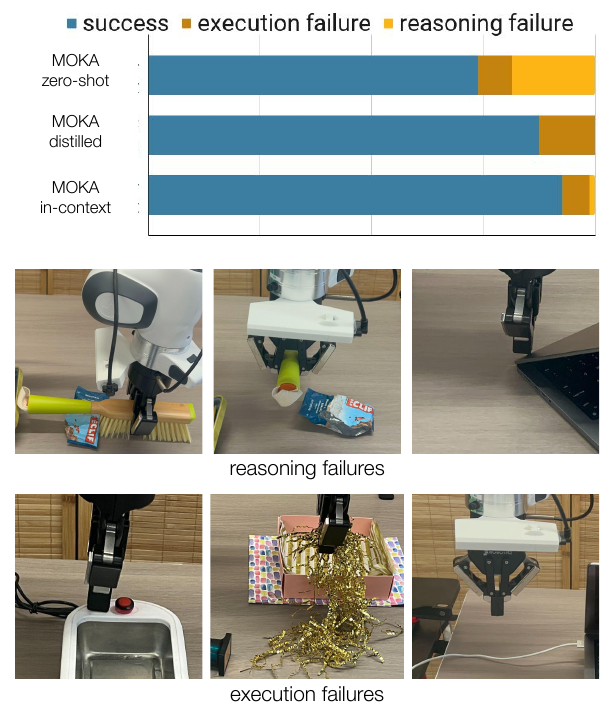} 
    \caption{\textbf{Failure analysis.} We break down the failure cases of the variants of our approach using zero-shot prediction, distilled policies, and in-context learning. We break down the failures into reasoning failures (caused by errors in the affordance prediction) and execution failures (caused by low-level motions). We demonstrate how policy distillation and in-context learning reduce failure cases.}
    \label{fig:failure_breakdown}
\end{figure}

\begin{table}[h!]
\begin{tabular}{ll}
                                       &  \\ \hline
\multicolumn{1}{l|}{\multirow{2}{*}{Table Wiping}} & 1. Move away the glasses \\ \cline{2-2} 
\multicolumn{1}{l|}{}                  &  2. Sweep the trash with the broom \\ \hline
\multicolumn{1}{l|}{\multirow{2}{*}{Watch Cleaning}} & 1. Put the watch into the ultrasound cleaner \\ \cline{2-2} 
\multicolumn{1}{l|}{}                  & 2. Press the power button \\ \hline
\multicolumn{1}{l|}{\multirow{2}{*}{Gift Preparation}} & 1. Put the golden filler in the gift box \\ \cline{2-2} 
\multicolumn{1}{l|}{}                  & 2. Put the perfume in the gift box \\ \hline
\multicolumn{1}{l|}{\multirow{2}{*}{Laptop Packing}} & 1. Unplug the cable\\ \cline{2-2} 
\multicolumn{1}{l|}{}                  &  2. Close the lid of the laptop\\ \hline
\end{tabular}
\caption{The subtask instructions of each of the testing tasks. Each of the testing tasks consists of two subtasks.}
\label{tab:task_desion}
\end{table}

\subsection{Experimental Setup}
\label{sec:experiments_setup}
We compare \MethodAcronym~with \textbf{Code-as-Policies}~\citep{liang2023code} and \textbf{VoxPoser}~\citep{huang2023voxposer}, two baselines that also enable zero-shot execution of open-world tasks:. Code-as-Policies provides a framework for language model-generated programs executed on robotic systems by prompting with code examples. For a fair comparison, we provide the two baselines with the task description in the code comments, with an additional 40 lines of code prompts providing example usage, as in the original implementation~\citep{liang2023code}. Similarly, VoxPoser~\citep{huang2023voxposer} also provides code examples to large language models to build a 3D voxel map of value functions. For VoxPoser, we reuse the example prompts and planning pipeline in the original implementation and use the same hyper-parameters to create the voxel value map. For both baselines, we only adapt the perception modules for fair comparisons while retaining the functionality of the other components.

We evaluate \MethodAcronym~in both zero-shot and in-context learning settings and refer to them as \textbf{\MethodAcronym~zero-shot} and \textbf{\MethodAcronym~in-context}, respectively. For the in-context learning setting, we provide two examples to GPT-4V with instructions and annotated images before querying information about the current task. The annotated images are collected in scenes with object and instruction variations. Some in-context prompt examples can be found in the Appendix.

\rebuttal{The successful trajectories generated by \MethodAcronym~can also serve as demonstration data for other learning-based methods. By simply training a model using the successful trajectories generated from \MethodAcronym, we can also ``distill'' the knowledge from a VLM to a learned policy. To achieve this goal, we employ a recent open-sourced robot foundation model Octo~\citep{octo_2023}, which is pre-trained on a diverse mix of 800K robot trajectories~\citep{padalkar2023open} and can be easily fine-tuned to new tasks. Octo~\citep{octo_2023} supports language instructions or goal images as task specifications and takes current observations to generate actions with a transformer-based diffusion policy architecture. It also provides code examples to easily perform fine-tuning. To construct our fine-tuning dataset, we collect 50 successful trajectories for each task with language annotations. We then adopt its latest base model checkpoint\footnote{https://huggingface.co/rail-berkeley/octo-base} and finetune the full model with the recipe provided in the official instruction\footnote{https://github.com/octo-models/octo}. The performance of fine-tuned can be found in Sec.~\ref{sec:experiments_quantatitive} with more implementation details in Appendix~\ref{sec:policy-distill}. Although \MethodAcronym~is a training-free method, it can be leveraged for data-driven approaches to provide training data for open-world tasks. We refer to this approach as \MethodAcronym-Distilled. }

\subsection{Quantitative Evaluation}
\label{sec:experiments_quantatitive}
Our quantitative evaluation results across 4 tasks are illustrated in TABLE~\ref{tab:experiment}. For each task, we report the number of successes out of 10 trials. As shown in the table, \MethodAcronym~achieves state-of-the-art performance at each subtask of the 4 tasks (total 8 subtasks), with consistent improvements using in-context learning. On most of the tasks, VoxPoser~\citep{huang2023voxposer} has similar performance with \MethodAcronym~(zero-shot), except for subtask 2 of table wiping (which is a tool-use task). Additionally, the task success rates can be sensitive to the resolution of the voxel map, which requires some hyperparameter tuning. Unlike the baselines, \MethodAcronym~can work well without example prompts or can achieve better performance with just two clean and intuitive example prompts.

We also observed that we can obtain an effective end-to-end policy by fine-tuning a pre-trained robot
 policy~\citep{octo_2023} using the successful trajectories generated by \MethodAcronym. This suggests that the generated data is of high quality and thus can be used as demonstration data for open-world tasks. The fact that distilling the successful \MethodAcronym~trials actually \emph{improves} the overall performance of the system suggests the feasibility of using \MethodAcronym~as a ``data generator'', where the zero-shot \MethodAcronym~method is used to bootstrap a continuous improvement system. This is an exciting direction to explore in future work.

\textbf{Failure breakdown}
We analyze the failure cases of \MethodAcronym.
Trajectories with wrong predictions from the GPT-4V are counted as reasoning failures. The following failure cases are illustrated in the top row of Fig.~\ref{fig:failure_breakdown}, including grasping the broom upside down due to confusing the grasp point with the function point, pointing the room in the wrong direction due to the wrong target angle, pressing the hinge of the laptop due to the wrong target point.
The trajectories with desired VLM prediction but fail to execute successfully are counted as execution failures. The following failure cases are illustrated in the bottom row of Fig.~\ref{fig:failure_breakdown}. From left to right, the failures include the gripper narrowly missing the button, the filler getting disassembled in the middle of the trajectory, and the cable slipping through the gripper fingers. 

As shown in the figure, both policy distillation and in-context learning reduce the total failures. 
\rebuttal{Using the distilled policy, the VLM is no longer part of the pipeline. Therefore, there are no reasoning failures in this case}.

\subsection{Qualitative Evaluation}
\label{sec:experiments_qulitative}

In Fig.~\ref{fig:example_results}, we provide six examples of the annotated marks, generated motions, and the unrolled trajectories of \MethodAcronym. In each of these examples, based on the annotated marks, \MethodAcronym~successfully selected the keypoints and waypoints for generating the desired motions. These examples demonstrate that our approach can be applied to various skills, including pressing, closing, rearranging, tool use, \etc. 

\begin{figure}[t]
    \centering
    \includegraphics[width=0.48\textwidth]{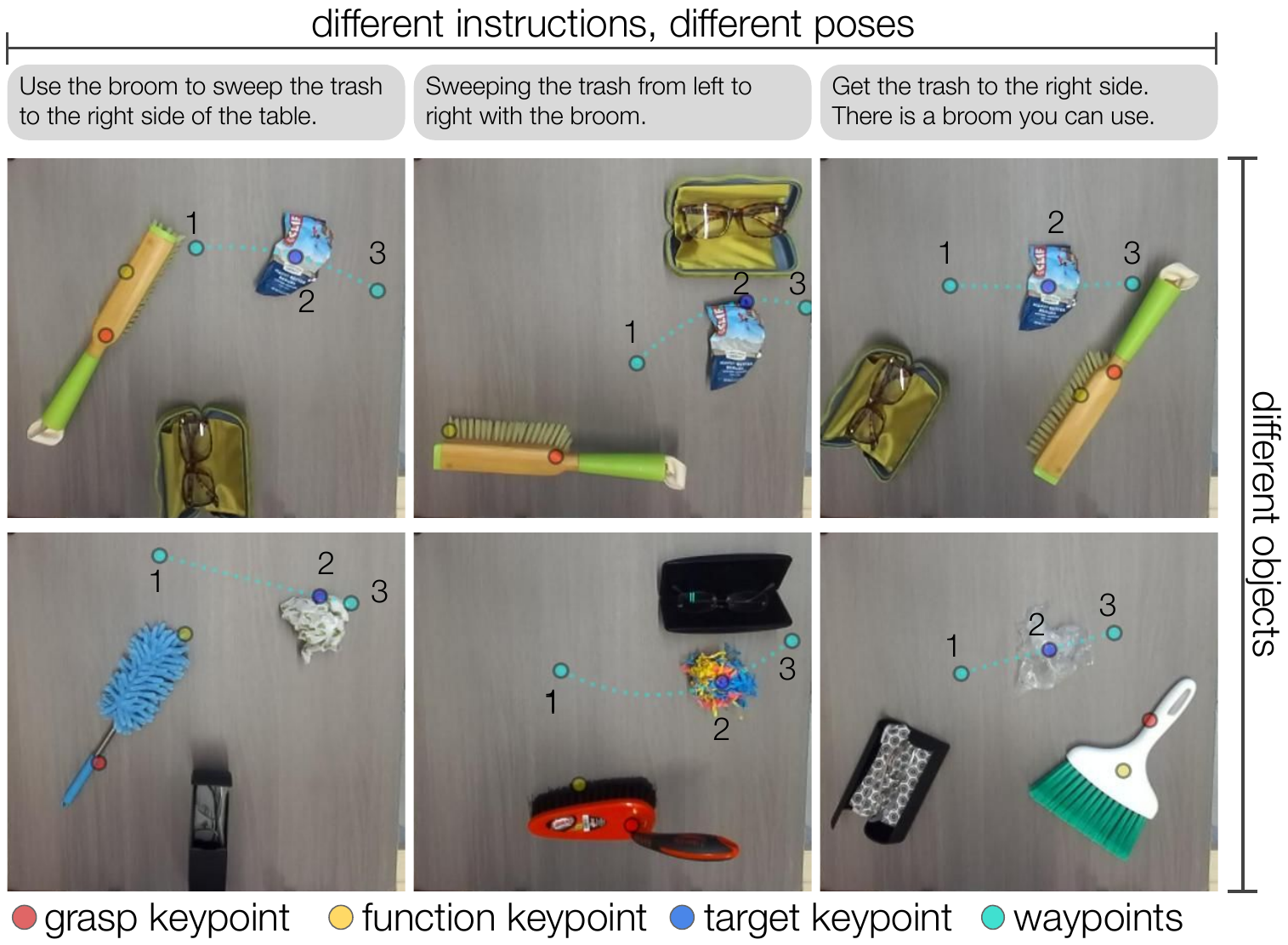} 
    \caption{\textbf{Robustness analysis.} We analyze \MethodAcronym's robustness with respect to various instructions, initial arrangements, and objects. Each column in the image uses the same language instruction and similar initial arrangements of objects. The two rows involve different objects. \rebuttal
    \MethodAcronym~ demonstrates to robustly predicts the keypoints and waypoints across a wide range of variations. More examples of other tasks are presented in the Appendix.}
    \label{fig:robustness}
\end{figure}

We analyze the robustness of \MethodAcronym~with respect to instruction variations, initialization variations, and object variations. Fig.~\ref{fig:robustness} provides a pictorial illustration of \MethodAcronym's predictions under different instructions and observations. Each image is queried with the language prompt on the top within the same column. The examples in the same column share similar initialization object positions and orientations. The example in the first row uses the same set of objects, while the second row involves objects of alternative geometries, colors, and materials. Some of these objects are also deformable or transparent. Fig.~\ref{fig:robustness} demonstrates consistent keypoint and waypoint predictions within rows and columns, indicating that \MethodAcronym~is robust to the changes of the language instructions, objects, and initial arrangements for the same task.

\section{Conclusion and Discussion}

In this paper, we propose \MethodAcronym, a simple and effective approach that leverages pre-trained VLMs for open-world robotic manipulation. 
By specifying motions with point-based affordance representations, we enable VLMs pre-trained on Internet-scale data to solve manipulation tasks in zero-shot and few-shot manners.
Using the proposed mark-based visual prompting technique, \MethodAcronym~converts affordance reasoning into a series of visual question-answering problems that the pre-trained VLMs can solve. 
Through experiments on a variety of table-top manipulation tasks, we demonstrate the effectiveness and robustness of \MethodAcronym. 
We also demonstrate that the trajectories generated by \MethodAcronym~can be further used to bootstrap the performance through in-context learning and policy distillation. 
As far as we know, \MethodAcronym~is the very first work that leverages visual prompting on pre-trained VLMs for open-world robot manipulation. 
We hope \MethodAcronym~will inspire future research towards leveraging vision-language models for open-world robotic control.

\MethodAcronym~is subject to limitations mostly due to the capabilities of existing VLMs and the current design of the affordance representations. 
The state-of-the-art VLMs still lack a profound understanding of 3D space, contact physics, and robotic control, which prevents our approach from generating more fine-grained motions in the $SE(3)$ space.
The high latency of querying VLMs also makes it challenging for the robot to perform dynamic manipulation and open-loop control.
Moreover, we would like to note that, while we demonstrate that a unified point-based affordance representation can be devised to support diverse tasks, the specific design of the representation in this work does not cover all useful skills that a robot can perform.
To apply \MethodAcronym~to more complex scenarios, \eg, bi-manual manipulation and whole-body control, we would need to extend the set of points and include additional attributes in this affordance representation. 
We believe these limitations will be addressed as more capable vision-language models and robotic foundation models are introduced in the future. 

\section*{Acknowledgement}
This research was supported by the AI Institute, AFOSR FA9550-22-1-0273, and the NSF under IIS-2246811. 
Fangchen Liu was supported in part by NSF under the NSF AI4OPT Center. 
We would like to thank Karl Pertsch, Oleg Rybkin, and Zheyuan Hu for providing feedback
on early drafts of the manuscript. 
We also want to thank Sudeep Dasari, Qiyang Li, Yide Shentu, and Homer Walke for valuable suggestions on the experiments and infrastructure.

\clearpage
\newpage

\appendices

\section{Experiment Details}
\subsection{Environment}

Our experiments are conducted in a real-world table-top manipulation environment, which involves a 7-DoF Franka Emika robot arm with a 2F-85 Robotiq gripper interacting with various objects on the table at 5Hz.
Our tabletop environment has two fixed ZED 2.0 cameras and one ZED mini wrist camera that can take RGBD images.

The \textit{top-down camera}, which is the primary camera used in this paper, provides RGB and the depth images for \MethodAcronym~and other baseline methods.

In addition, we set up the front camera and the wrist camera for more complete observation to distill the collected robot experiences to the learned Octo policy~\cite{octo_2023}. The action space of the robot is 7-dimensional, consisting of the 6-DoF end-effector twist defined in the $SE(3)$ space with an additional dimension of gripper position.
Each task can consist of multiple stages, each terminates after 100 steps.

\subsection{Task Design}
\label{sec:full-task}
We design various table-top manipulation tasks with a diverse set of daily objects. TABLE~\ref{tab:task_desion_full} shows the language description of the full list of tasks.

\begin{table}[h!]
\begin{tabular}{ll}
                                       &  \\ \hline
\multicolumn{1}{l|}{\multirow{2}{*}{Table Wiping}} & 1. Move the glasses into the glasses case \\ \cline{2-2} 
\multicolumn{1}{l|}{}                  &  2. Sweep the trash with the broom \\ \hline
\multicolumn{1}{l|}{\multirow{2}{*}{Watch Cleaning}} & 1. Put the watch into the ultrasound cleaner \\ \cline{2-2} 
\multicolumn{1}{l|}{}                  & 2. Press the power button \\ \hline
\multicolumn{1}{l|}{\multirow{2}{*}{Gift Preparation}} & 1. Put the golden filler in the gift box \\ \cline{2-2} 
\multicolumn{1}{l|}{}                  & 2. Put the perfume in the gift box \\ \hline
\multicolumn{1}{l|}{\multirow{2}{*}{Laptop Packing}} & 1. Unplug the cable\\ \cline{2-2} 
\multicolumn{1}{l|}{}                  &  2. Close the lid of the laptop\\ \hline

\multicolumn{1}{l|}{\multirow{2}{*}{Fur Removing}} & 1. Grasp the fur remover\\ \cline{2-2} 
\multicolumn{1}{l|}{}                  &  2. Sweep the fur on the sweater with the remover\\ \hline
\multicolumn{1}{l|}{\multirow{2}{*}{Drawer Closing}} & 1. Close the drawer\\ \cline{2-2} 
\multicolumn{1}{l|}{}                  &  \\ \hline
\multicolumn{1}{l|}{\multirow{2}{*}{Scissor Handing}} & 1. Grasp the scissor \\ \cline{2-2} 
\multicolumn{1}{l|}{}                  &  2. Hand the scissor to a human\\ \hline
\multicolumn{1}{l|}{\multirow{2}{*}{Flower Arrangement}} & 1. Grasp the pink roses on the table\\ \cline{2-2} 
\multicolumn{1}{l|}{}                  &  2. Insert the roses into the vase\\ \hline
\end{tabular}
\caption{The language description of all the proposed tasks. Each of the tasks consists of two stages except for the drawer closing task.}
\label{tab:task_desion_full}
\end{table}

We provide the comparative evaluation on the top four tasks (Table Wiping, Laptop Packing, Gift Preparation and Untrasound Cleaning) in the main paper, with additional results of \MethodAcronym~on other four tasks in our supplementary video.

\subsection{Success Checking}
To count the failure modes and collect successful trajectories, we first query VLM with task instruction and obtain the point-based motion prediction. If the prediction is correct, it will be executed on the robot. Otherwise, it will be counted as the VLM reasoning failure as mentioned in Sec.~\ref{sec:experiments_quantatitive}. After executing the motion prediction from VLM, the human expert will manually check if the task succeeds or not. If the task is not successfully finished, it will be counted as an execution failure. All the successful trajectories will be saved as demonstrations for policy distillation.

\section{Implementation Details}
\label{sec:implementation}

We introduce the overview of \MethodAcronym~in Alg.~\ref{alg:overview}, and the implementation details of each component in the following sections.
\begin{algorithm}[ht] 
\caption{\MethodAcronym~Pipeline} 
\begin{algorithmic}[1]\label{alg:overview}
  \STATE \textbf{Input:} Vision-language Model $\mathcal{M}$, Task instruction $l$, text prompt for high-level reasoning $p_{high}$, text prompt for low-level reasoning $p_{low}$, initial observation $s_0$
  \STATE Query $\mathcal{M}$ for high-level task reasoning, obtain $y_{high} = \mathcal{M}([p_{high}, l, s_0])$ which decompose the task into $N$ subtasks.
  \FOR{subtask $k=0 \cdots N-1$}
  \STATE Get observation $s_k$ from the top-down camera
  \STATE Propose keypoint and waypoint candidates and get annotated image $f(s_k)$
  \STATE Query $\mathcal{M}$ for low-level motion reasoning, obtain $y_{low}^k = \mathcal{M}([p_{low}, y_{high}^k, f(s_k)])$
  \STATE Execute $y_{low}^k$ on the real robot
  \ENDFOR
\end{algorithmic}
\end{algorithm}

\subsection{High-Level Task Reasoning}
Given the initial observation image $s_0$ and the language task description $l$, we first query the VLM $\mathcal{M}$ with the language instruction to produce the decomposed subtask, which contains the structured
information for the $K$ subtasks, including descriptions of objects and desired motion. We provide the prompt for high-level reasoning in TABLE~\ref{tab:high-level}. 

\begin{table*}[htbp]
\begin{mdframed}%

The input request contains: 
\begin{itemize}
    \item A string describes the multi-stage task.
    \item An image of the current table-top environment captured from a top-down camera.
\end{itemize}
The output response is a list of dictionaries in the JSON form. Each dictionary specifies the information of a subtask, following the correct order of executing the subtasks to solve the input task. Each dictionary contains these fields: 
\begin{itemize}
    \item \textbf{instruction}: A string to describe the subtask in natural language forms.
    \item \textbf{object$\_$grasped}: A string to describe the name of the object that the robot gripper will hold in hand while executing the task (\eg,, the object to be picked or used as a tool to interact with other objects). This field can be empty if there is no such object involved in this subtask.
    \item \textbf{object$\_$unattached}: A string to describe the name of the object that the robot gripper will interact with directly or via another object without holding it in hand (\eg,, the object to be touched by the tool, the target object where ``object$\_$grasped'' will be moved onto). This field can be empty if there is no such object involved in this subtask.
    \item \textbf{motion$\_$direction}: A string to describe the direction of the robot gripper motion while performing the task (\eg,, ``from right to left'', ``backward'', ``downward'').
\end{itemize}
\end{mdframed}
\caption{The high-level reasoning prompt for all the tasks. It will decompose a multi-stage task into subtasks. The specified output fields can be used for later low-level reasoning stage.}
\label{tab:high-level}
\end{table*}

The response contains fields ``object$\_$grasped'', ``object$\_$unattached'' and ``motion$\_$direction'', which will be used for later stages in our pipeline, such as keypoint proposal and motion prediction.

\subsection{Point-based Affordance Proposal}
\label{sec:keypoint-implementation}
After obtaining the high-level reasoning results, we can know the objects involved in each subtask. Since VLM cannot directly generate keypoints and waypoints, we need to propose some candidate points and let VLM select corresponding points through a visual question-answering way.

\textbf{Keypoint proposal.} We leverage GroundedSAM~\citep{ren2024grounded} to extract segmentation masks conditioned on a text prompt, which is designed to be a string of object names involved in the current subtask (\eg, ``trash, broom''). Given an image of current observation and such a text prompt about the involved object names, we first employ GroundingDINO~\citep{liu2023grounding} to generate bounding boxes for the objects by conditioning on the text prompt. Later, the annotated boxes given by GroundingDINO~\citep{liu2023grounding} serve as the box prompts for SAM~\citep{kirillov2023segment} to generate corresponding segmentation masks for the objects. We define the keypoints of each object to be a set of boundary keypoints with a center keypoint. To extract them, we sample $K$ points on the contour of each object using farthest point sampling~\citep{qi2017pointnet}, and also include the geometric mean point of the object segmentation mask. We then plot these $K+1$ keypoints on the 2D image as our keypoint proposals to the VLM.

\begin{figure}[t]
    \centering
    \includegraphics[width=0.45\textwidth]{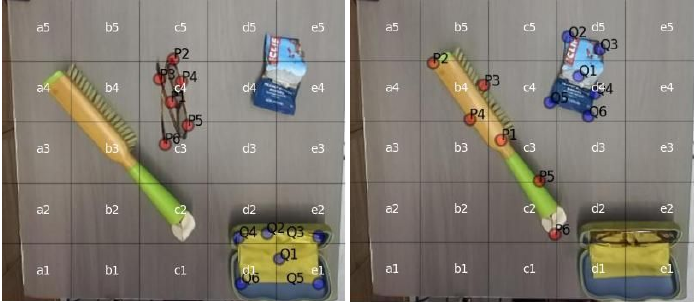} 
    \caption{The affordance proposal for the table wiping task in stage 1 and stage 2. The keypoints are plotted on corresponding objects for different stages based on the high-level reasoning results. The annotated grid cells are used for VLM to select the position of potential waypoints.}
    \label{fig:keypoint-waypoint}
\end{figure}

\textbf{Waypoint proposal.} Unlike keypoints, waypoints are often the points in the free space that are not attached to any objects. To perform waypoint selection, we evenly divide the full image into $5\times5$ grid tiles,  which mark the column as \textit{a, b, c, d, e} from left to right and rows as \textit{1, 2, 3, 4, 5} from bottom to top, as shown in Fig.~\ref{fig:keypoint-waypoint}. The waypoints will be sampled from the predicted tile from VLM.

\subsection{Low-Level Motion Reasoning}
\label{sec:low-level}
After annotating the observation image using object keypoints and grid cells as described in the above section, we can query VLM about low-level motion with annotated image and text prompts. Firstly, we describe the text and image inputs to the VLM using prompt~\ref{tab: input-description}.

\begin{table*}[htbp]
\begin{mdframed}
Please describe the robot gripper's motion to solve the task by selecting keypoints and waypoints.

The input request contains:

    \begin{itemize}
        \item The task information with these fields:
        \begin{itemize}
            \item \textbf{instruction}: The task in natural language forms.
            \item \textbf{object$\_$grasped}: The object that the robot gripper will hold in hand while executing the task.
            \item \textbf{object$\_$unattached}: The object that the robot gripper will interact with either directly or via another object without holding it in hand. 
            \item \textbf{motion$\_$direction}: The motion direction of the robot gripper or the in-hand object while performing the task.
        \end{itemize}
        \item An image of the current table-top environment captured from a top-down camera, annotated with a set of visual marks:
        \begin{itemize}
            \item \textbf{candidate keypoints on object$\_$grasped}: Red dots marked as $P[i]$ on the image.
            \item \textbf{candidate keypoints on object$\_$unattached}: Blue dots marked as $Q[i]$ on the image.
            \item \textbf{grid for waypoints}: Grid lines that uniformly divide the images into tiles. The grid equally divides the image into columns marked as $a, b, c, d, e$ from left to right and rows marked as 1, 2, 3, 4, 5 from bottom to top. 
        \end{itemize}
    \end{itemize}
\end{mdframed}
\caption{The description about image and text inputs in the low-level motion reasoning stage.}
\label{tab: input-description}
\end{table*}

Then we first explain the definition of keypoints and waypoints using prompt~\ref{tab:explanation}.

\begin{table*}
\begin{mdframed}
The motion consists of an optional grasping phase and a manipulation phase, specified by \textbf{grasp$\_$keypoint}, \textbf{function$\_$keypoint}, \textbf{target$\_$keypoint}, \textbf{pre$\_$contact$\_$waypoint}, and \textbf{post$\_$contact$\_$waypoint}. 

The definitions of these points are:
    \begin{itemize}
        \item \textbf{grasp$\_$keypoint}: The point on ``object$\_$grasped'' indicates the part where the robot gripper should hold.
        \item \textbf{function$\_$keypoint}: The point on ``object$\_$grasped'' indicates the part that will make contact with ``object$\_$unattached''.
        \item \textbf{target$\_$keypoint}: If the task is pick-and-place, this is the location where ``object$\_$grasped'' will be moved to. Otherwise, this is the point on ``object$\_$unattached'' indicating the part that will be contacted by ``function$\_$keypoint''. 
        \item \textbf{pre$\_$contact$\_$waypoint}: The waypoint in the free space that the functional point moves to before making contact with the ``target$\_$keypoint''.
        \item \textbf{post$\_$contact$\_$waypoint}: The waypoint in the free space that the functional point moves to after making contact with the ``target$\_$keypoint''.
    \end{itemize}
 \end{mdframed}
 \caption{The explanation about the defination of keypoints and waypoints.}
 \label{tab:explanation}
\end{table*}
After that, we can specify the output format and further explain the role of each field using prompt~\ref{tab:motion-output}. We can also provide some step-by-step guidance about how the reasoning procedure should be done (similar to chain-of-thought prompting~\citep{wei2022chain}). 
\begin{table*}
\begin{mdframed}
The response should be a dictionary in JSON form, which contains:
\begin{itemize}
    \item \textbf{grasp$\_$keypoint}: Selected from candidate keypoints marked as $P[i]$ on the image. This will be empty if and only if object$\_$grasped is empty.
    \item \textbf{function$\_$keypoint}: Selected from candidate keypoints marked as $P[i]$ on the image. This will be empty if and only if object$\_$grasped or object$\_$unattached is empty.
    \item \textbf{target$\_$keypoint}: Selected from keypoint candidates marked as $Q[i]$ on the image. This will be empty if and only if object$\_$unattached is empty. 
    \item  \textbf{pre$\_$contact$\_$tile}: The tile that the pre-contact waypoint should be in. This is selected from candidate tiles marked on the image.
    \item  \textbf{post$\_$contact$\_$tile}: The tile that post-contact waypoint should be in. This is selected from candidate tiles marked on the image.
    \item \textbf{pre$\_$contact$\_$height}: The height of pre-contact waypoint as one of the two options ``same'' or ``above'' (same or higher than the height of making contact with target keypoint).
    \item \textbf{post$\_$contact$\_$height}: The height of post-contact waypoints as one of the two options ``same'' or ``above''.
    \item \textbf{target$\_$angle}: Describe how the object should be oriented during this motion in terms of the axis pointing from the grasping point to the function point. 
    
\end{itemize}
Think about this problem step by step and explain the reasoning steps. First, choose grasp$\_$keypoint, function$\_$keypoint, and target$\_$keypoint on the correct parts of the objects. Next, describe which tile the target$\_$keypoint is located in. Then choose pre$\_$contact$\_$tile, post$\_$contact$\_$tile, pre$\_$contact$\_$height, post$\_$contact$\_$height such that the resultant motion from pre-contact waypoint to target keypoint, then to post-contact waypoint in 3D follows the ''motion$\_$direction'' input. Remember that the columns are marked as 'a', 'b', 'c', 'd', 'e' from left to right, and the rows are marked as 1, 2, 3, 4, 5 from bottom to top. 
\end{mdframed}
\caption{The output format of the low-level motion reasoning, including some further explanations.}
\label{tab:motion-output}
\end{table*}

\subsection{Motion Execution}
We can decompose a manipulation subtask into an \textbf{optional} grasping phase and a manipulation phase. For some tasks like tool using, the robot needs to first grasp an object before making contact with another object. For other tasks where the robot can directly interact with target objects (\eg, pushing, pressing button), the ``grasp$\_$keypoint'' field will be empty. For such tasks, we will skip the execution of grasping phase.

\textbf{Grasping phase.}
We first sample 30 antipodal 4-DoF grasp proposals based on the primary camera's depth image, which is cropped by the bounding box of \textbf{object$\_$grasped}. Here, we use the same antipodal grasp proposal method used in DexNet 2.0~\citep{mahler2017dex}. We then lift the VLM-predicted grasp point from 2D image to the robot frame based on the primary camera's intrinsic and extrinsic parameters. After that, we can select the nearest grasp proposal based on the lifted grasp point, and execute the 4-DoF grasp on the robot.

\textbf{Manipulation phase.}
After obtaining the pre-contact tile and post-contact tile, we first sample pre-contact waypoint and post-contact waypoint from the predicted tiles. Since the waypoints are in the free space, its 3D position cannot be determined given its 2D projection on the image. So we also assign the VLM-predicted height to them. We only consider the cases where the waypoints are in same height with target point (\eg, pushing), or above the target point (\eg, placing) in most common table top manipulation scenarios. 

We then sequentially move the function keypoint to the pre-contact waypoint, target keypoint and post-contact waypoint to perform a contact motion. 

After both phases are successfully done, the robot need to get the motion prediction for the next subtask. In order to obtain a clean and non-occuluded image to perform low-level reasoning, we first move the robot to the neural pose, get an observation image from the top-down camera, and then resume the robot. Subsequently, the robot will execute the predicted motion until the multi-stage task is finished.

\rebuttal{
\subsection{Policy Distillation}
\label{sec:policy-distill}
We employ the base checkpoint of Octo~\cite{octo_2023} and perform full-model finetuning using their official instruction. The default architecture of Octo is a transformer-based diffusion policy, which is conditioned on a task discription or goal image through a task tokenizer along with tokenized pixel observations from multiple cameras, and predicts actions through a diffusion head. The diffusion head consists of a 3-layer MLP with a hidden dimension of 256 using a standard DDPM objective~\citep{ho2020denoising}.

We mostly reuse the hyperparameters as in Octo~\cite{octo_2023}. For customized hyper-parameters we used, please refer to TABLE~\ref{tab:octo}. }
\begin{table}[th]
\centering
\begin{tabular}{cc}
\toprule
Hyperparameter
     &  Value \\
     \midrule
  Learning Rate   & 3e-4\\
  Warmup Steps & 1000\\
  Weight Decay & 0.01\\
  Gradient Clip Threshold & 0.5\\
  Batch Size & 256\\
  Total Gradient Steps & 200000\\
  \bottomrule
\end{tabular}
\caption{Hyperparameters used during fine-tuning.}
\label{tab:octo}
\end{table}

\subsection{Evaluation}
We evaluate \MethodAcronym~in both zero-shot and in-context learning settings. For the zero-shot setting, we keep all the above prompts \textbf{unchanged} but only change the language task description (\eg, ``sweep the garbage'', ``pick up the perfume'', ``press the button''). 

\begin{figure*}[htbp]
    \centering
    \includegraphics[width=\textwidth]{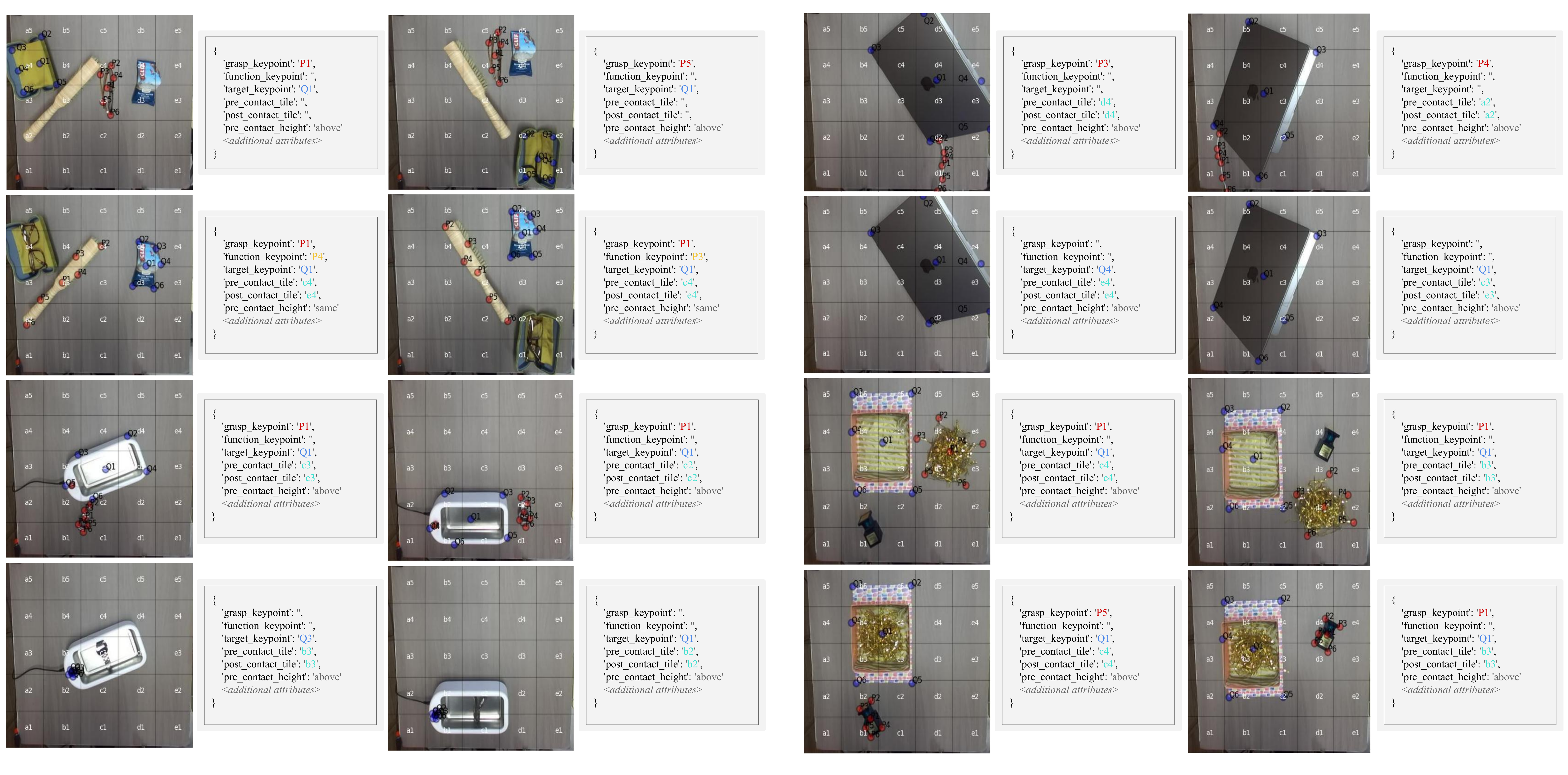} 
    \caption{The in-context examples used in \MethodAcronym~which are collected under object pose variations. }
    \label{fig:context}
\end{figure*}

For in-context learning settings, we first collect two examples from VLM's successful predictions in different scenes with scene and object variations. As shown in Fig.~\ref{fig:context}, \MethodAcronym~can be improved from such simple and intuitive in-context examples, without intricate prompt engineering. 

\section{Additional Results}
\label{sec:additional_results}
\subsection{Ablation Study}
We design the following ablative studies to understand the effectiveness of different design options in \MethodAcronym. 

\begin{itemize}
    \item \MethodAcronym~w/o hierarchy: we can skip the high-level task reasoning but directly ask for low-level motion reasoning from GPT-4V. Here all the objects in the scene will be annotated with keypoints, and VLM is queried to generate point-based affordance directly from the annotated image.
    \item \MethodAcronym~w/o description of points: we can remove the description of the definition of keypoints and waypoints in Tab.~\ref{tab:explanation}. 
    \item \MethodAcronym~w/o chain-of-thought prompting: we can remove the step-by-step guidance at the last paragraph in the prompt in Tab.~\ref{tab:motion-output}.
\end{itemize}

\begin{figure*}[htbp]
    \centering
    \includegraphics[width=\textwidth]{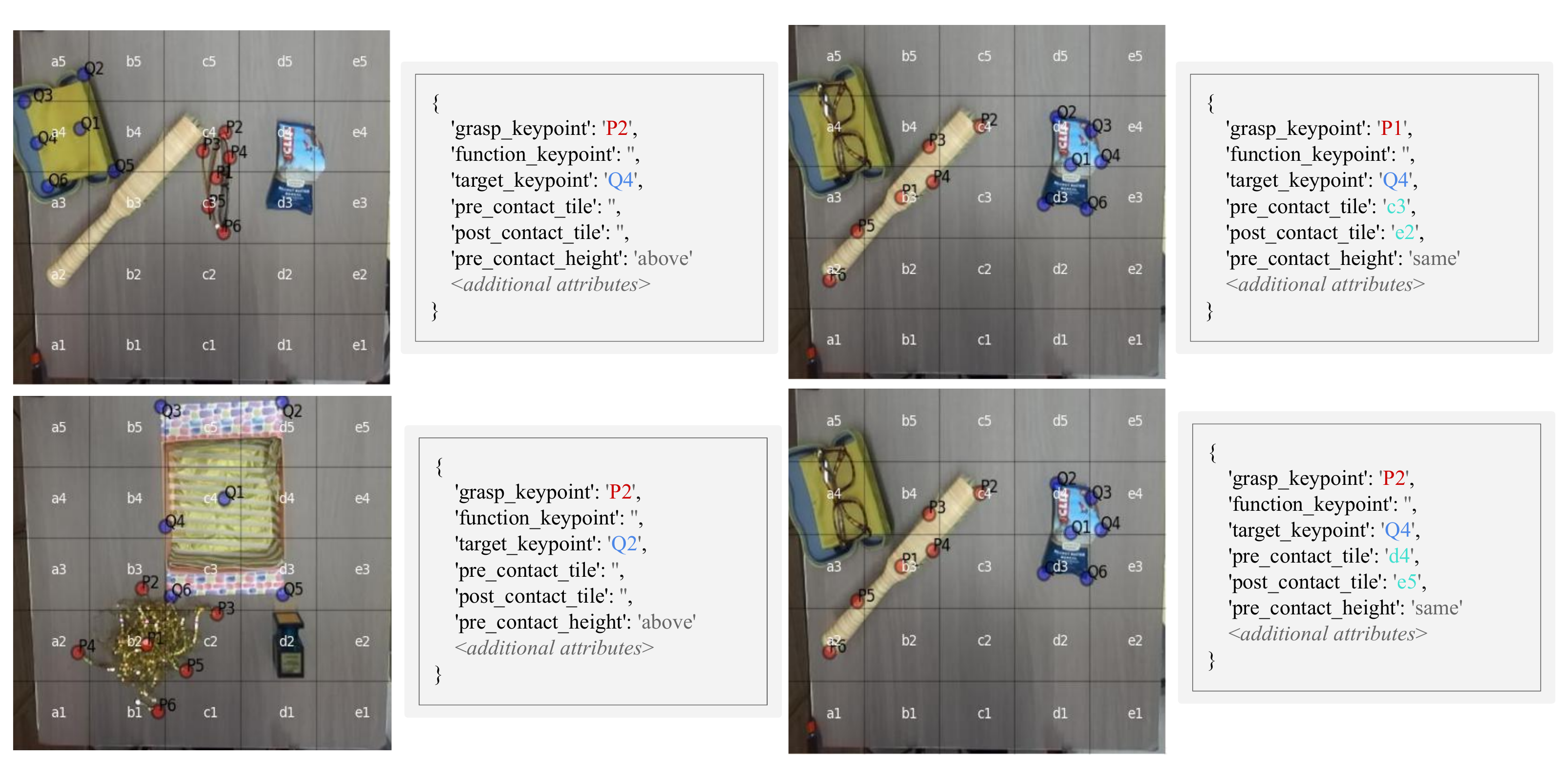} 
    \caption{Qualitative results of ablation studies. Without the keypoint description or chain-of-thought prompting, the model can make mistakes in keypoint prediction (left column) and waypoint prediction (right column).}
    \label{fig:ablation}
\end{figure*}

The results of our ablative studies are illustrated in Tab.~\ref{tab:ablations}. For each task, we report the number of reasoning successes rate out of 10 trials. The results demonstrate that our method can obtain consistent improvements from all the above prompting designs. Specifically, \textbf{\MethodAcronym~w/o hierarchy} decreases the performance by a large margin, which indicates the importance of subtask decomposition. \textbf{\MethodAcronym~w/o keypoint description} and \textbf{\MethodAcronym~w/o CoT} can preserve most of the performance, but still worse than \MethodAcronym~on most of the tasks. As illustrated in Fig.~\ref{fig:ablation}, VLM can make various mistakes in terms of keypoint and waypoint predictions. Specifically, for \MethodAcronym~w/o hierarchy, the VLM can make mistakes about subtask ordering, which can cause a complete failure of the task.

\begin{table*}[htbp]
    \centering
    \begin{tabular}{lcccccccc}  
    \toprule
    & \multicolumn{2}{c}{Table Wiping} & \multicolumn{2}{c}{Watch Cleaning} & \multicolumn{2}{c}{Gift Preparation} & \multicolumn{2}{c}{Laptop Packing} \\
    \cmidrule(lr){2-3} \cmidrule(lr){4-5} \cmidrule(lr){6-7} \cmidrule(lr){8-9}\\
    & Subtask I & Subtask II & Subtask I & Subtask II & Subtask I & Subtask II & Subtask I & Subtask II  \\
     & \\
    \midrule
    MOKA                   & 0.7           & 0.7 & 0.8 & 1.0 & 1.0 & 0.8 & 0.6 & 0.8 \\
    \hline
    MOKA w/o hierarchy     & 0.1           & 0.1 & 0.2 & 0.1 & 0.2 & 0.2 & 0.0 & 0.1 \\
    MOKA w/o keypoint description   & 0.5  & 0.6 & 0.7 & 0.8 & 0.9 & 0.6 & 0.4 & 0.8 \\
    MOKA w/o CoT    & 0.5           & 0.4 & 0.6 & 0.8 & 0.9 & 0.6 & 0.4 & 0.6 \\
    \bottomrule
    \end{tabular}
    \caption{Ablation studies on different prompt designs. Across 4 tasks, each consists of 2 subtasks, \MethodAcronym~consistently benefits from the three prompt designs.}
    \label{tab:ablations}
\end{table*}

\subsection{Additional Robustness Analysis}
\begin{figure*}[t]
    \centering
    \begin{minipage}{0.45\textwidth}
        \centering
        \includegraphics[width=\textwidth]{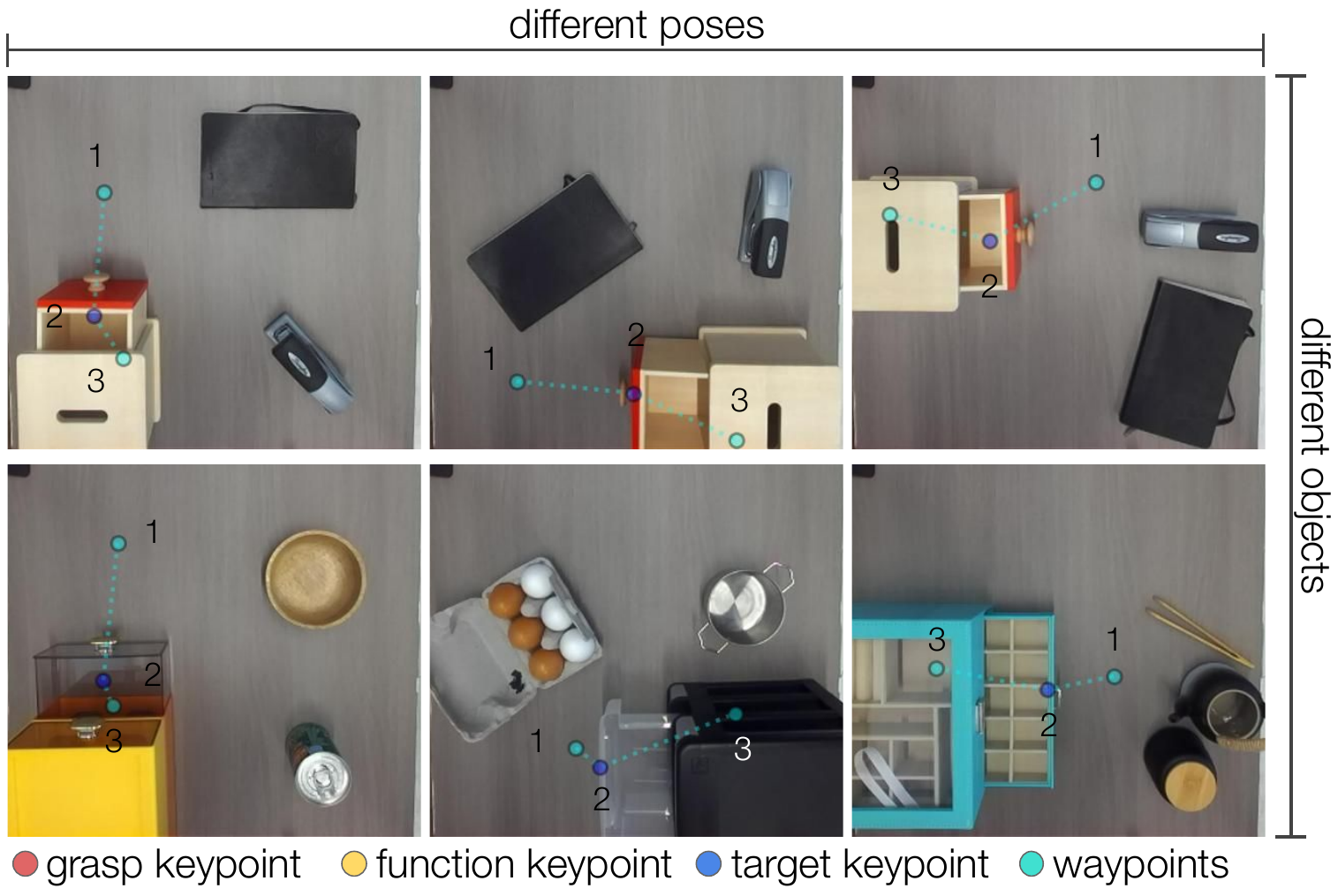}
        \caption{\rebuttal{Robustness analysis for the drawer closing task.}}
        \label{fig:robustness_drawer}
    \end{minipage}\hfill
    \begin{minipage}{0.45\textwidth}
        \centering
        \includegraphics[width=\textwidth]{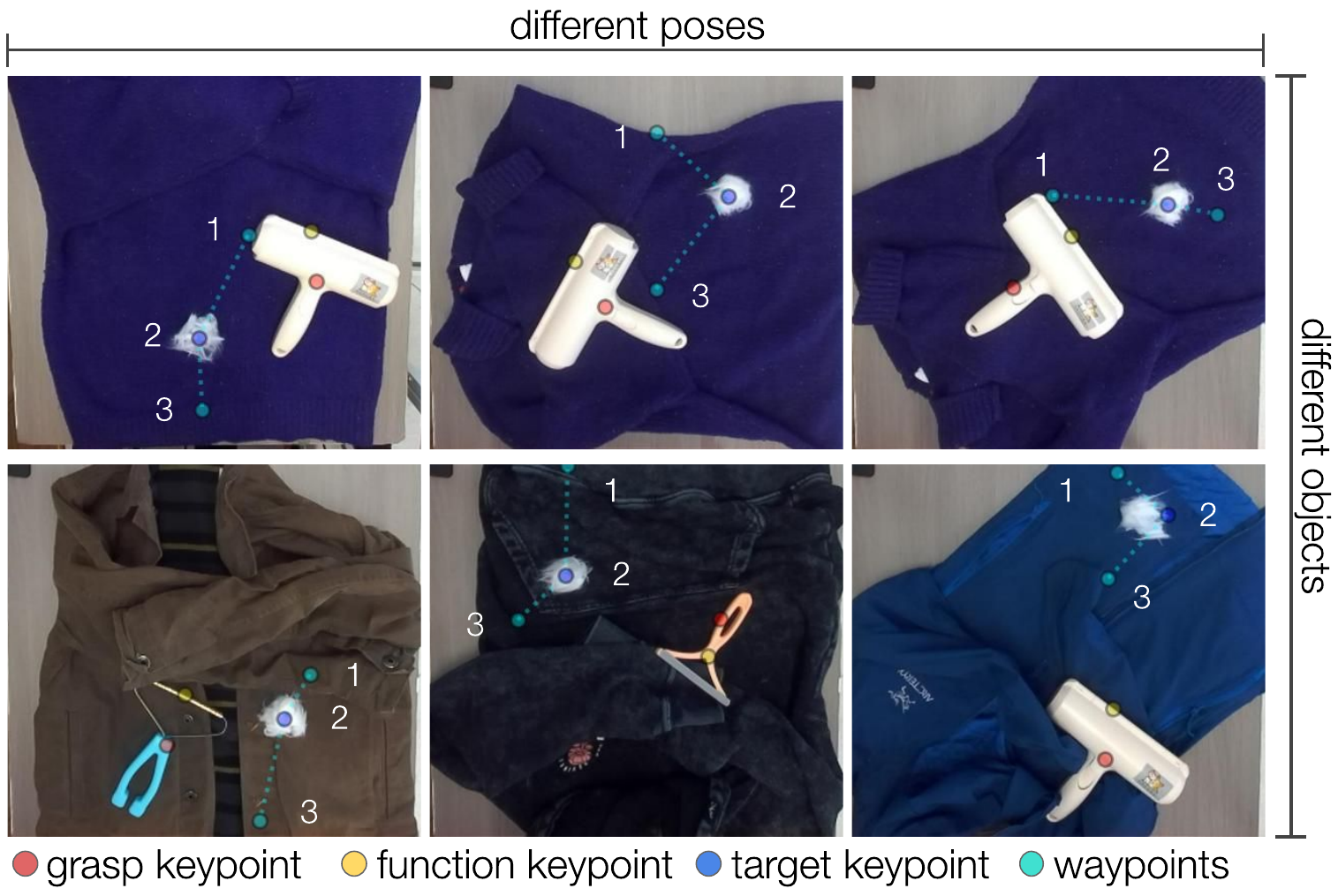}
        \caption{\rebuttal{Robustness analysis for the fur removal task.}}
        \label{fig:robustness_fur_removal}
    \end{minipage}
\end{figure*}

We provide additional results for robustness analysis. 
The predicted affordances for the drawer closing and fur removal tasks with various objects and initial poses are shown in Fig.~\ref{fig:robustness_drawer} and Fig.~\ref{fig:robustness_fur_removal}. 
In these two figures, each column in the image uses similar initial arrangements of objects, while the two rows use different objects. For the drawer closing tasks, MOKA can consistently predict the waypoints and target points following the correct closing motion regardless of the poses and appearances of drawers and other distractors. For the fur removal tasks, MOKA can predict all the keypoints correctly with a feasible motion in different environments.

\end{document}